\documentclass[conference]{IEEEtran}

\ifCLASSOPTIONcompsoc
  \usepackage[nocompress]{cite}
\else
  \usepackage{cite}
\fi

\ifCLASSINFOpdf
\usepackage[pdftex]{graphicx}
\else
\fi

\usepackage[float]{}

\usepackage{algorithm}
\usepackage{algpseudocode}
\usepackage{textcomp}
\usepackage[english]{babel}
\usepackage[utf8x]{inputenc}
\usepackage{amsmath}
\usepackage{amssymb}
\usepackage{graphicx}
\usepackage[export]{adjustbox}
\usepackage{array,multirow}
\usepackage[table]{xcolor}
\usepackage{url}

\usepackage{listings}
\usepackage{cleveref}

\usepackage[colorinlistoftodos]{todonotes}
\graphicspath{ {figures/} }
\makeatletter
\DeclareTextCommandDefault{\textleftarrow}{\mbox{$\m@th\leftarrow$}}
\makeatother

\begin{document}

\title{``Did You Hear That?''\\ Learning to Play Video Games from Audio Cues}

\IEEEoverridecommandlockouts
\IEEEpubid{\begin{minipage}{\textwidth}\ \\[12pt]
978-1-7281-1884-0/19/\$31.00 \copyright 2019 IEEE
\end{minipage}}

\author{
\IEEEauthorblockN{Raluca D. Gaina}
\IEEEauthorblockA{\textit{Game AI Research Group}\\
Queen Mary University of London, UK\\
r.d.gaina@qmul.ac.uk}
\and
\IEEEauthorblockN{Matthew Stephenson}
\IEEEauthorblockA{\textit{Department of Data Science and Knowledge Engineering}\\
Maasricht University, NL\\
matthew.stephenson@maastrichtuniversity.nl}
}

\IEEEtitleabstractindextext{%
\begin{abstract}
Game-playing AI research has focused for a long time on learning to play video games from visual input or symbolic information. However, humans benefit from a wider array of sensors which we utilise in order to navigate the world around us. In particular, sounds and music are key to how many of us perceive the world and influence the decisions we make. In this paper, we present initial experiments on game-playing agents learning to play video games solely from audio cues. We expand the Video Game Description Language to allow for audio specification, and the General Video Game AI framework to provide new audio games and an API for learning agents to make use of audio observations. We analyse the games and the audio game design process, include initial results with simple Q-Learning agents, and encourage further research in this area.
\end{abstract}
}

\maketitle

\IEEEdisplaynontitleabstractindextext

\section{Introduction} \label{sec:intro}

Sound and music have long been an important aspect of video game development and play \cite{Bartle2003}. Not only can audio greatly influence our engagement and emotional investment in a game \cite{backgroundmusic}, but it can also provide important environmental information or gameplay cues \cite{zenouda2012}. Sounds within games can be used to alert the player to a nearby hazard (especially when in darkness), inform them they collected an item, or provide clues for solving certain puzzles.
This additional sensory output is different from traditionally visual information, and allows for many new gameplay possibilities. 

Certain games rely heavily on audio to create an immersive atmosphere, particularly horror games \cite{horrorsound}, and would likely lose much of their impact without it. Other games can sometimes require the player to listen to and understand certain sounds to progress, such as the numerous music and audio based puzzles in point-and-click adventure games (Myst, The Secret of Monkey Island, Machinarium, etc.). Even when not essential for the player to proceed, many games use sound to inform the player about useful non-visual information, such as alarm systems in many stealth games (Thief, Far Cry, Alien: Isolation, etc.) or enemy positions in first person shooters (Overwatch, Call of Duty, Battlefield, etc.)
Without the ability to process audio input, we would likely be unable to play many of these games effectively. Some expert human players are also able to play games exclusively based on audio input even where on the surface visuals would appear essential, such as visually impaired players competing in fighting game tournaments \cite{blind} or speedrunners attempting to finish games such as Mario, Zelda or Punch Out while blindfolded.

While such examples are specific to the domain of video games, detecting and understanding certain sounds can be vitally important in a variety of other real-world scenarios. In such cases, current machine learning approaches that do not consider audio input as a factor in their decisions, could be seriously hindered. One topical example could be for a self-driving car \cite{car}. Ambulances, fire engines and other emergency response vehicles typically use loud sirens and bright lights to alert traffic to pull over. It is often the case when driving that you can hear the sound produced by these vehicles long before you can see them. Most human drivers would be able to detect an approaching emergency response vehicle by sound and react appropriately. A self-driving car relying only on visual input would only react once it saw the vehicle itself. It is very possible to imagine many similar scenarios where the ability to listen to and interpret audio would be an important skill to possess. Returning to video games, any situation where there is an audible sound for something that is not yet visible on screen could be of benefit to an agent.

The remainder of this paper is structured as follows. Section~\ref{sec:bg} gives a brief background on the analysis of audio in games and the framework used in this work. Section~\ref{sec:gvgai-sound} describes the expansion of the General Video Game AI framework to include audio cues. Section~\ref{sec:exp} discusses initial experiments, while Section~\ref{sec:end} concludes with future work.




\section{Background}\label{sec:bg}

\subsection{Audio analysis in games}

AI has previously been used in several ways to analyse the effects and meaning of game audio. TagATune is a game that involves annotating sounds and music \cite{law2007tagatune}. The collection of categorised audio clips that an application like this can produce might allow AI to process and learn the likely effects of different output sounds (e.g. predict whether a particular sound indicates that the player is being healed or damaged). This idea of understanding the intended meaning of different sounds is also related to the field of AI-based audio stenography \cite{zamani2009artificial}, which focuses on interpreting noisy or corrupt audio messages.
Several AI approaches for providing a categorical understanding of spoken dialogue systems have also been proposed \cite{potamianos2005adaptive}, including the use of deep reinforcement learning \cite{weisz2018sample}. This has several applications within video games, such as understanding human speech and generating suitable response options for NPCs. 
The topics of procedural audio generation \cite{Edwards2011} and sound synthesis 
are also highly relevant, as sounds produced by certain actions often depend on their consequences. For example, collecting a coin would likely yield an entirely different sound to that of killing an enemy. An interesting connection can be done to the work by Lopes et al.~\cite{lopes2015sonancia} and, more generally, the area of soundscape generation for games: where the \textit{Sonacia} systems decides what sounds should be played in a procedurally generated level, our AI could learn to interpret them and react accordingly.







\subsection{General Video Game AI Framework}

We chose to develop an audio game-playing API within an existing framework in order to make use of existing agents (and visual-based games) as a starting point for the project. The General Video Game AI framework (GVGAI)~\cite{perez2018gvgai} is a Java framework which contains a large (and continuously expanding) set of games written in the Video Game Description Language (VGDL), which propose varied challenges for game-playing agents, from navigation to puzzles to fast-reactionary problems. In addition to the large collection of games, we consider an important benefit the fact that new games can easily be created in VGDL, as well as varying existing ones to obtain a potentially infinite supply of games. The Learning Track in the GVGAI competition proposes the challenge of developing general learning agents based on either visuals (an image of the game state can be provided) or symbolic information. In Section~\ref{sec:gvgai-sound} we describe the expansion of the framework by adding the possibility of including audio signals in VGDL, as well as processing these appropriately in the Java framework and sending correct observations to the agents.


\section{Audio Games in GVGAI}\label{sec:gvgai-sound}

We integrated audio with VGDL in order to provide sound properties for games in two of the definition sets:
\begin{itemize}
    \item \texttt{\textbf{SpriteSet}}: Each of the sprites can have audio files associated with them in the format \texttt{audio=move:filename1;use:filename2}. Audio signals are integrated into 3 functionalities: 
    \begin{itemize}
        \item Sprite movement: audio plays on each sprite move.
        \item \textit{USE}-type\footnote{In GVGAI, \textit{USE} actions are used for avatars as the fifth legal action in some games, which can have various effects, from spawning new sprites of different types to jumping or activating some avatar-specific properties} sprite action: audio plays whenever the sprite applies a \textit{USE}-type action and it can be applied to spawn points as well as avatars, e.g. "play chime sound each time a new enemy is spawned".
        \item Beacon sprite: audio plays at every game tick (volume based on the proximity to the player's avatar).
    \end{itemize}
    Several or none of the options can be defined per sprite. 
    \item \texttt{\textbf{InteractionSet:}} Each of the interactions defined in GVGAI can have audio files associated with them in the format \texttt{audio=filename}. The \textit{SoundManager} class is called to play the defined sound (if any) at the beginning of each interaction, i.e. when 2 sprites that have a defined effect overlap. We have also added a new interaction option, \textit{playSound}, which only plays the given sound, without any other effects taking place.
\end{itemize}

The \textit{SoundManager} features several easy-to-use methods for sound management, including playing, pausing or restarting sounds. It uses a pre-defined path and file extension (thus only the name of the file within the path, without the extension, should be included in VGDL). In the current version, only \textit{.wav} files are supported.

In addition to the VGDL integration, a new AI agent API and game-running options have been introduced as well. Audio-ony game-players should extend the \textit{AudioPlayer} class and implement the required methods. They receive \textit{AudioStateObservation} objects, which restrict observations to sound only: the agents are only provided with an array of \textit{AudioObservation} objects, ordered by the proximity to the avatar (closest observations will be first). Each \textit{AudioObservation} object includes details of its proximity to the avatar, as well as information about the \textit{.wav} file associated with the observation for quick processing: bytes, fingerprint, normalized amplitudes and spectogram. The \textit{.wav} file itself is also supplied, to allow for use of custom libraries in audio processing.


The codebase is publicly available on GitHub\footnote{\url{https://github.com/rdgain/GVGAI-Sound}}.





\subsection{Audio Games Set}\label{sec:games}

This section describes the 3 games we have included with the initial version of the framework. 

\texttt{\textbf{ALIENS}}. The player controls a spaceship which can move left and right, as well as shoot the incoming aliens, with the goal of killing all aliens. Bases can protect the player from incombing alien bombs, but can also be destroyed by player missiles. This is an audio adaptation of the original game in the GVGAI framework. Audio signals play each time the avatar shoots, aliens drop bombs, bombs or missiles kill bases or aliens, and when the avatar hits the edge of the play area.

\texttt{\textbf{LABYRINTH}}. A simple maze navigation game, where each level comprises of paths in-between impassable walls. This game is simplified from the GVGAI version with the removal of traps. Audio signals play each time the avatar bumps into walls, as well as every tick for the exit, a beacon sprite. 

\texttt{\textbf{BLOODSHED}}. A fighting game in which the avatar can move left or right and wave its sword in the direction it is currently facing. This game is adapted from the 2015 Samtupy Productions audio game ``Bloodshed, release the pain''\footnote{\url{https://www.audiogames.net/}}. Audio signals play each time the avatar bumps into the edge of the play area, when they wave their sword, when they hit an enemy fighter or when the player is hit, with different sounds depending on the direction of the attack (left or right).

\section{Experiments and Discussion}\label{sec:exp}

This section describes some simple proof-of-concept experiments and analysis of the framework and games included. To this extent, we have adapted the sample Q-Learning method from the GVGAI Learning track~\cite{perez2018gvgai} to work with the new API. In particular, we now describe a state $S_t$ in terms of audio observations received by the agent, $O_{t0}, O_{t1} ... O_{tn}$ corresponding to any audio signals triggered at game tick $t$. $Q(S_t,a_t)$ is then updated depending on the reward received according to Equation~\ref{eq:ql}.

\begin{equation}\label{eq:ql}
\footnotesize{Q(s_t,a_t)\leftarrow Q(s_t,a_t)+\alpha [ r_{t+1} + \gamma \max_a Q(s_{t+1},a)-Q(s_t,a_t)]}
\end{equation}

As the game score is not available as part of the observation space, we replace the reward perceived by the agent with a heuristic evaluation of the state $S_t$. The audio observations are checked against the agent's knowledge base (a simple HashMap in this implementation) and the given weights are averaged for a final value of state $S_t$. The reward $r$ is then calculated as the difference in value to the previous state, i.e. $r = V(S_t) - V(S_{t-1})$. The knowledge base is updated at the end of each game played using a game-play trace consisting of a list of $(S_t, a_t)$ state-action pairs and the final game result (1 for win, -1 for loss), discounted such that the weights for pairs at the beginning of the game are affected most.

In the simplest case, the HashMap only stores a mapping from sound type to weight. The agent would then learn that some sounds are desirable in a game, while others are not.
The agent then picks actions which are thought to lead to positively-rewarded sounds. An intuitive extension from this is the addition of observation intensity in the knowledge base: the information would then be more specific in avoiding dangerous situations occurring close to the avatar. We call the first agent \textit{Q-KBS} and the second \textit{Q-KBI} and compare both against a random agent on the games described in Section~\ref{sec:games}. 

The agents are unable to consistently solve Bloodshed or Labyrinth. All agents achieve 0\% win rate in Bloodshed, but Q-KBI is able to score most average points in 1000 runs of the first level ($1.552$, compared to $1.314$ for Q-KBS and only $1$ for the random player). In Aliens, both learning agents are able to outperform random (49\% win rate for Q-KBI and 37\% win rate for Q-KBS compared to 32\% random), although this is far from the 100\% win rate achieved by planning agents. Our preliminary results do indicate that the learning show promise and should be further developed and analysed.

\begin{figure}[!t]
    \centering
    \includegraphics[width=\columnwidth]{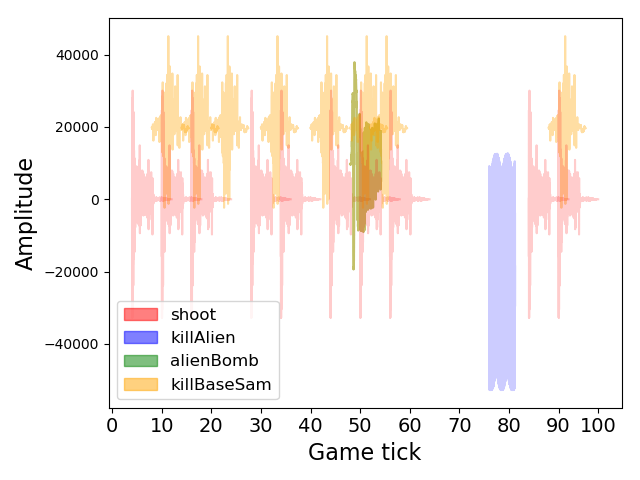}
    \caption{Audio observations in the first 100 ticks of the first level in the game \textbf{Aliens}, played by a random agent. The agent mostly observes feedback on its successful shooting actions (in red), which hit and destroy the protective bases (in orange). One of the avatar's bullets travels for longer (hence the pause in shooting between ticks 60 and 85) and eventually hits an alien (in blue). One of the aliens also drops a bomb (in green), which doesn't hit anything. Proximity to the avatar is not taken into account in this example, all sounds are played at full volume.}
    \label{fig:alienSounds}
\end{figure}

\subsection{Designing Audio Games}\label{sec:design}

In this context, we can further highlight a parallel made in previous literature~\cite{liapis2019orchestrating}, between designing games and orchestrating large musical pieces to which several different instruments contribute. Here we can see each sprite or interaction as a possible instrument which need to play the right notes at the right time. Overlaying too much input would overload the human processing capacity and such an audio game would become unplayable. Adding too little input would also make the game impossible to play, as there would not be sufficient feedback for the player's attempted actions in order for the player to figure out how the game works, or for them to be able to map out the game world.

Therefore, an in-depth analysis of audio games and the optimal level of input would prove interesting and useful. A first type of analysis proposed is a visual representation of audio observations received by the agents while playing the game. In Figure~\ref{fig:alienSounds}, we observe that the random agent takes into account actions which don't actually have an effect on the game, most apparent in the section between ticks 60-85, where the agent keeps trying to shoot, although this action has no effect within the rules of the game (only 1 player bullet can be in play at a time). Using audio feedback, a learning agent could realise that only once the bullet they shot hits an object are they allowed to shoot again; it could then focus its efforts on strategic positioning and planning further ahead.

A second type of analysis we consider is automatic pruning of non-essential or misleading audio signals. After 100 runs in the first level of the game Labyrinth, the agents' knowledge bases are shown in Table~\ref{tab:kb}, where ``bump'' is the audio signal for hitting walls and ``exit'' is the the exit point beacon. We can consider sounds with weights close to 0 non-essential, or highly fluctuating weights during learning as misleading.

\begin{figure}[!t]
    \centering
    \includegraphics[trim={0 5cm 0 0},clip,width=\columnwidth]{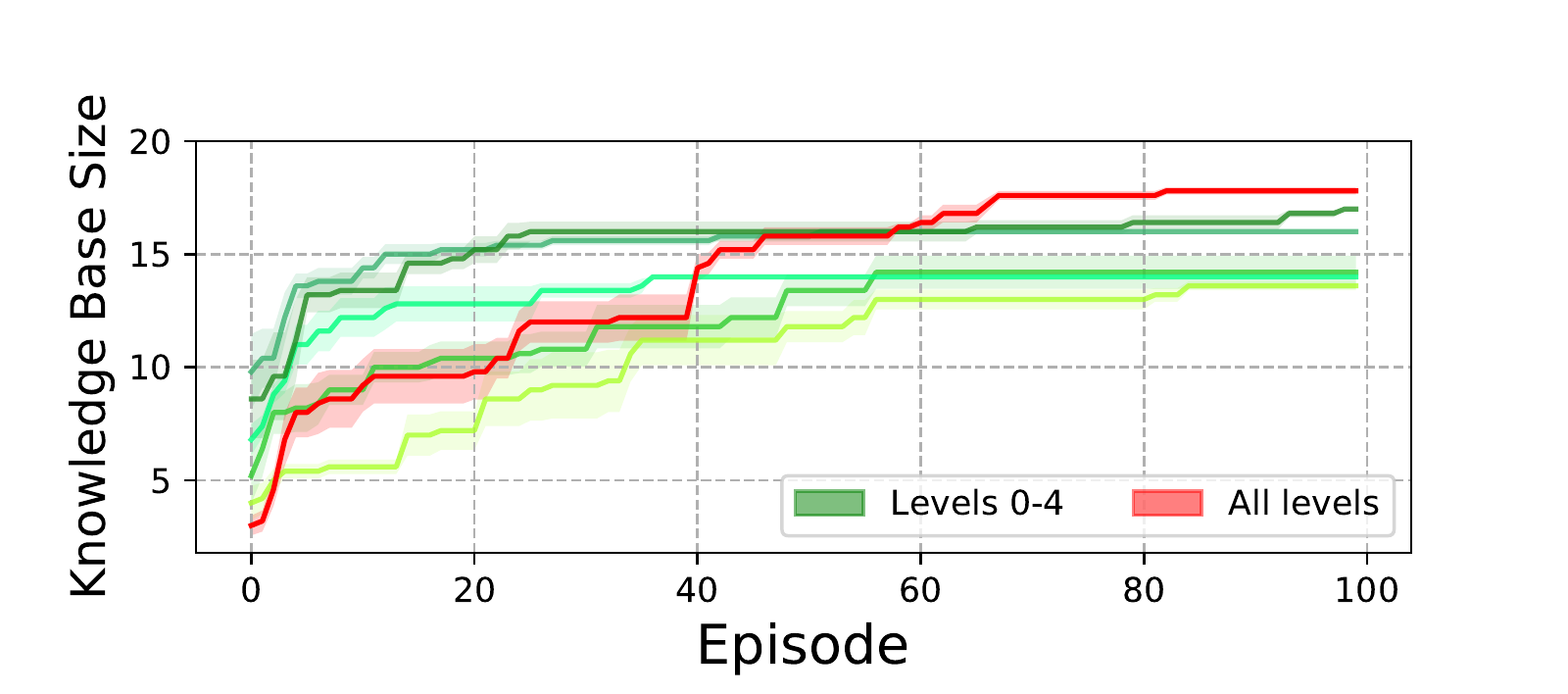}
    \caption{Knowledge base increase observed for Q-KBI agent in Labyrinth.}
    \label{fig:kbp}
\end{figure}

Regarding performance, the Q-KBS agent appears to learn the beacon sound is bad.
However, the agent loses most games (0.08\% win rate), so it associates the constant beacon sound with its consistent loss, while the bump sound receives less penalty as it occurs less. Given more time, we hypothesise this player would learn to avoid hitting walls, but it would struggle to find the exit. 
In contrast, the Q-KBI agent learns positive weights for being close to the exit and negative weights for being far away and for hitting walls (see Table~\ref{tab:kb}, where a list of \{sound intensity; learned associated weight) pairs is depicted\}. Given more time, we hypothesise this player would learn to consistently find the exit. Figure~\ref{fig:kbp} shows this agent's knowledge base size progression in the game Labyrinth (training on individual levels in green, and on all levels in red over 100 runs each). We can observe that in some levels the agent is able to move around more and learn more about the environment, while it appears beneficial to use more than 1 level for developing the knowledge. Using better learning algorithms and better exploration policies would help this agent in learning more efficiently to increase its performance.

\section{Conclusions} \label{sec:end}

In this paper we describe work in progress regarding an interesting direction of game-playing AI research: learning to play video games from audio cues only. We highlight that current state-of-the-art techniques rely either on visuals or symbolic information to interpret their environment, whereas humans benefit from the processing of many other types of sensor inputs. Sounds and music are key elements in games, which not only affect player experience, but gameplay itself in certain scenarios. We also introduce an extension of the General Video Game AI framework to support audio in games and audio observations. Simple Q-Learning agents were suggested to have promise in such environments.

There are many research directions opened by this work. A question that might arise is why not simply include observations of events or sprites behaviour? 
In partial observability scenarios (i.e. Starcraft), important events may take place outside of the player's vision range which are often signalled through sound effects. These sounds can then be analysed in more detail in order to create an appropriate response: while often intuitive for humans, machines can make use of sentiment analysis research~\cite{jain2019systematic}, for example, to identify what specific sounds might mean. 
Another interesting line of future work regarding audio analysis would be looking into exactly how removing certain sounds affects the agent's performance, in the context of having an agent proficient enough to be able to solve the games given enough information.
The knowledge and abilities learned by such high-skilled audio game-playing agents can be used together with other methods to maximise sensor usage for superior performance.

\begin{table}[!t]
\centering
\caption{Knowledge Base recorded by 2 agents after 100 runs on the first level of the audio game Labyrinth}\label{tab:kb}
\begin{tabular}{ccl}
\hline
\textbf{Agent} & \textbf{Sound} & \textbf{Recorded Weights}  \\
\hline
Q-KBS  & bump & -0.2412              \\
Q-KBS  & exit & -0.7194             \\
\hline
Q-KBI  & bump & [(1.00;-0.820)]            \\
Q-KBI  & exit & [(0.07;-0.990), (0.12;-0.108), (0.18;-0.959),\\
&&(0.25;-0.985), (0.33;0.914), (0.50;0.985)]            \\
\hline
\end{tabular}
\end{table}


Audio design in games also raises some important challenges when it comes to inclusivity and accessibility \cite{gameaccess}. People who may be partially or completely blind rely exclusively on audio, as well as some minor haptic feedback, to play a large number of video games effectively \cite{Yuan2009}. Including audio as well as visual information within a game can make completing it much more plausible for visually impaired players. Additionally, individuals with hearing difficulties 
would find it hard to play games that are heavily reliant on sound \cite{musiceffect}. Intelligent agents can help to evaluate games for individuals with disabilities: if an agent is able to successfully play a game using only audio or visual input, then this could help validate the game for the corresponding player demographics.






\bibliographystyle{IEEEtran}
\bibliography{IEEEabrv,main}

\begin{thebibliography}{10}
\providecommand{\url}[1]{#1}
\csname url@samestyle\endcsname
\providecommand{\newblock}{\relax}
\providecommand{\bibinfo}[2]{#2}
\providecommand{\BIBentrySTDinterwordspacing}{\spaceskip=0pt\relax}
\providecommand{\BIBentryALTinterwordstretchfactor}{4}
\providecommand{\BIBentryALTinterwordspacing}{\spaceskip=\fontdimen2\font plus
\BIBentryALTinterwordstretchfactor\fontdimen3\font minus
  \fontdimen4\font\relax}
\providecommand{\BIBforeignlanguage}[2]{{%
\expandafter\ifx\csname l@#1\endcsname\relax
\typeout{** WARNING: IEEEtran.bst: No hyphenation pattern has been}%
\typeout{** loaded for the language `#1'. Using the pattern for}%
\typeout{** the default language instead.}%
\else
\language=\csname l@#1\endcsname
\fi
#2}}
\providecommand{\BIBdecl}{\relax}
\BIBdecl

\bibitem{Bartle2003}
R.~Bartle, \emph{{Designing Virtual Worlds}}.\hskip 1em plus 0.5em minus
  0.4em\relax New Riders Games, 2003.

\bibitem{backgroundmusic}
J.~Zhang Xiaoqing~Fu, ``{The Influence of Background Music of Video Games on
  Immersion},'' \emph{PPRS}, vol.~05, 2015.

\bibitem{zenouda2012}
H.~Z{\'e}nouda, ``{New Musical Organology : the Audio-Games},'' in
  \emph{{MISSI'12 - Int. Conf. on Multimedia \& Network Info. Systems}}, 2012.

\bibitem{horrorsound}
G.~Roux-Girard, ``Listening to fear: A study of sound in horror computer
  games,'' \emph{{Game Sound Technology and Player Interaction: Concepts and
  Developments}}, pp. 192--212, 2011.

\bibitem{blind}
D.~O'Keefe, ``{The Blind Masters of Fighting Games},''
  \url{https://tinyurl.com/gaming-blind}, 2018, [Online; accessed 14-May-2019].

\bibitem{car}
M.~Bojarski \emph{et~al.}, ``{End to End Learning for Self-Driving Cars},''
  \emph{CoRR}, vol. abs/1604.07316, 2016.

\bibitem{law2007tagatune}
E.~L. Law \emph{et~al.}, ``{TagATune: A Game for Music and Sound Annotation},''
  in \emph{ISMIR}, vol.~3, 2007, p.~2.

\bibitem{zamani2009artificial}
M.~Zamani, H.~Taherdoost, A.~A. Manaf, R.~B. Ahmad, and A.~M. Zeki, ``{An
  Artificial-Intelligence-Based Approach for Audio Steganography},''
  \emph{MASAUM Journal of Open Probelms in Science and Engineering (MJOPSE)},
  vol.~1, no.~1, pp. 64--68, 2009.

\bibitem{potamianos2005adaptive}
A.~Potamianos, S.~Narayanan, and G.~Riccardi, ``{Adaptive categorical
  understanding for spoken dialogue systems},'' \emph{IEEE Transactions on
  Speech and Audio Processing}, vol.~13, no.~3, pp. 321--329, 2005.

\bibitem{weisz2018sample}
G.~Weisz, P.~Budzianowski, P.-H. Su, and M.~Gasic, ``{Sample Efficient Deep
  Reinforcement Learning for Dialogue Systems with Large Action Spaces},''
  \emph{IEEE/ACM Transactions on Audio, Speech and Language Processing
  (TASLP)}, vol.~26, no.~11, pp. 2083--2097, 2018.

\bibitem{Edwards2011}
M.~Edwards, ``{Algorithmic Composition: Computational Thinking in Music},''
  \emph{Commun. ACM}, vol.~54, no.~7, pp. 58--67, 2011.

\bibitem{lopes2015sonancia}
P.~Lopes, A.~Liapis, and G.~N. Yannakakis, ``Sonancia: Sonification of
  procedurally generated game levels,'' in \emph{Proceedings of the 1st
  computational creativity and games workshop}, 2015.

\bibitem{perez2018gvgai}
D.~Perez \emph{et~al.}, ``{General Video Game AI: a Multi-Track Framework for
  Evaluating Agents, Games and Content Generation Algorithms},'' \emph{IEEE
  Transactions on Games}, 2019.

\bibitem{liapis2019orchestrating}
A.~Liapis \emph{et~al.}, ``{Orchestrating Game Generation},'' \emph{IEEE
  Transactions on Games}, vol.~11, no.~1, pp. 48--68, 2019.

\bibitem{jain2019systematic}
S.~K. Jain and P.~Singh, ``{Systematic Survey on Sentiment Analysis},'' in
  \emph{2018 First International Conference on Secure Cyber Computing and
  Communication (ICSCCC)}, 2019, pp. 561--565.

\bibitem{gameaccess}
B.~Yuan, e.~folmer, and F.~Harris, ``{Game accessibility: A survey},''
  \emph{Universal Access in the Information Society}, vol.~10, pp. 81--100,
  2011.

\bibitem{Yuan2009}
B.~Yuan, ``{Towards Generalized Accessibility of Video Games for the Visually
  Impaired},'' Ph.D. dissertation, Uni. of Nevada, Reno, 2009.

\bibitem{musiceffect}
K.~F. Hansen and R.~Hiraga, ``{The Effects of Musical Experience and Hearing
  Loss on Solving an Audio-Based Gaming Task},'' \emph{Applied Sciences},
  vol.~7, no.~12, 2017.

\end{thebibliography}

\end{document}